\documentclass[sigconf,authordraft]{format}

\AtBeginDocument{%
  \providecommand\BibTeX{{%
    \normalfont B\kern-0.5em{\scshape i\kern-0.25em b}\kern-0.8em\TeX}}}
\pdfoutput=1
\renewcommand\footnotetextcopyrightpermission[1]{}
\makeatletter
\def\ACM@mk@linecount{}
\def\ACM@linecountL{}
\def\ACM@linecountR{}
\makeatother

\usepackage{mathrsfs}
\usepackage{times}
\usepackage{soul}
\usepackage{url}
\usepackage{subfigure}
\usepackage[utf8]{inputenc}
\usepackage{caption}
\usepackage{graphicx}
\usepackage{amsmath}
\usepackage{amsthm}
\usepackage{booktabs}
\usepackage{algorithm}
\usepackage{algorithmic}
\usepackage[switch]{lineno}
\usepackage{multirow, multicol}
\usepackage{amsfonts}
\usepackage{enumitem}
\usepackage{color}
\usepackage{bm}

\newcommand{\eat}[1]{}
\newcommand{\eg}{{\em e.g., }}     
\newcommand{\ie}{{\em i.e., }}      

\AtBeginDocument{%
  \providecommand\BibTeX{{%
    Bib\TeX}}}

\begin{document}

\title{Noisy Node Classification\\ by Bi-level Optimization based  Multi-teacher Distillation}

\author{Yujing Liu}
\orcid{0009-0005-9229-5188}
\affiliation{%
  \institution{Guangxi Key Lab of Multisource Information Mining and Security, Guangxi Normal University}
  \city{Guilin}
  \country{China}
  \postcode{541004}
}
\email{liuyujingcn@gmail.com}

\author{Zongqian Wu}
\affiliation{%
  \institution{School of Computer Science and Engineering, University of Electronic Science and Technology of China}
  \city{Chengdu}
  \country{China}
  \postcode{611731}}
\email{wkzongqianwu@gmail.com}

\author{Zhengyu Lu}
\affiliation{%
  \institution{Guangxi Key Lab of Multisource Information Mining and Security, Guangxi Normal University}
  \city{Guilin}
  \country{China}
  \postcode{541004}
}
\email{airmanlu@foxmail.com}

\author{Ci Nie}
\affiliation{%
  \institution{Guangxi Key Lab of Multisource Information Mining and Security, Guangxi Normal University}
  \city{Guilin}
  \country{China}
  \postcode{541004}
}
\email{nieci2024@gmail.com}

\author{Guoqiu Wen}
\affiliation{%
  \institution{Guangxi Key Lab of Multisource Information Mining and Security, Guangxi Normal University}
  \city{Guilin}
  \country{China}
  \postcode{541004}
}
\email{wenguoqiu2008@163.com}

\author{Ping Hu}
\affiliation{%
  \institution{School of Computer Science and Engineering, University of Electronic Science and Technology of China}
  \city{Chengdu}
  \country{China}
  \postcode{610000}
}
\email{chinahuping@gmail.com}

\author{Xiaofeng Zhu}
\affiliation{%
  \institution{Guangxi Key Lab of Multisource Information Mining and Security, Guangxi Normal University}
  \city{Guilin}
  \country{China}
  \postcode{541004}
}
\email{seanzhuxf@gmail.com}


\begin{abstract}
Previous graph neural networks (GNNs) usually assume that the graph data is with clean labels for representation learning, but it is not true in real applications. In this paper, we propose a new multi-teacher distillation method based on bi-level optimization (namely BO-NNC), to conduct noisy node classification on the graph data. 
Specifically, we first employ multiple self-supervised learning methods to train diverse teacher models, and then aggregate their predictions through a teacher weight matrix. Furthermore, we design a new bi-level optimization strategy to dynamically adjust the teacher weight matrix based on the training progress of the student model. Finally, we  design a label improvement module to improve the label quality. 
Extensive experimental results on real datasets show that our method achieves the best results compared to state-of-the-art methods. \textit{The source code is listed in Supplementary Materials.}
\end{abstract}

\ccsdesc[500]{Computing methodologies~Neural networks}

\keywords{Graph convolutional networks, Noisy labels, Bi-level Optimization,  Semi-supervision node Classification}



\maketitle

\section{Introduction}

Node classification usually relies on correct supervised information to predict unlabelled nodes on the graph data.
However, real-world graph data often contain incorrect supervised information, \ie noisy labels, which easily misleads the training model and  in turn degrades the node classification performance. To address this issue, noisy node classification (NNC) has  been widely proposed to train machine learning models (\eg graph neural networks (GNNs)) on the graph data with noisy labels\footnote{\textbf{Related Work} is listed in Appendix A.} \cite{li2021unified}, so NNC has garnered growing interest in practical applications \cite{dai2021say,guo2019attention}.

A number of methods have been proposed to handle noisy labels. For example, NRGNN \cite{dai2021nrgnn} trains a pseudo-label miner to select pseudo-labels for augmenting the supervision information. UnionNET \cite{li2021unified} first trains a GNN using the original labels and then performs label correction based on the node embeddings learned by the obtained GNN. However, these methods excessively depend on the performance of individual model, and their effectiveness can significantly decrease as the noise rate increases.
For instance, in the case of high noise rates, NRGNN will select incorrect pseudo-labels and UnionNET will contaminate correct labels with label correction. 
One of the key reasons is that previous methods design a single model only to handle with noisy label to result in limited robustness.

To alleviate the above issue, the multi-teacher distillation method is proposed to transfer knowledge from multiple models to deal with noisy labels. For instance, MTS-GNN \cite{liu2023multi} uses models saved from earlier iterations to guide model training and label correction for later iterations. However, previous multi-teacher distillation methods still have limitations to be addressed. 
First, the constructed multi-teacher models across nearby iterations have few difference, and thus resulting in the lack of diversity.
Second, previous methods do not consider the complementary information among teacher models and could not fully exploit the knowledge of teacher models. 

\begin{figure*}[t!]
    \centering
    \includegraphics[width=1\textwidth,]{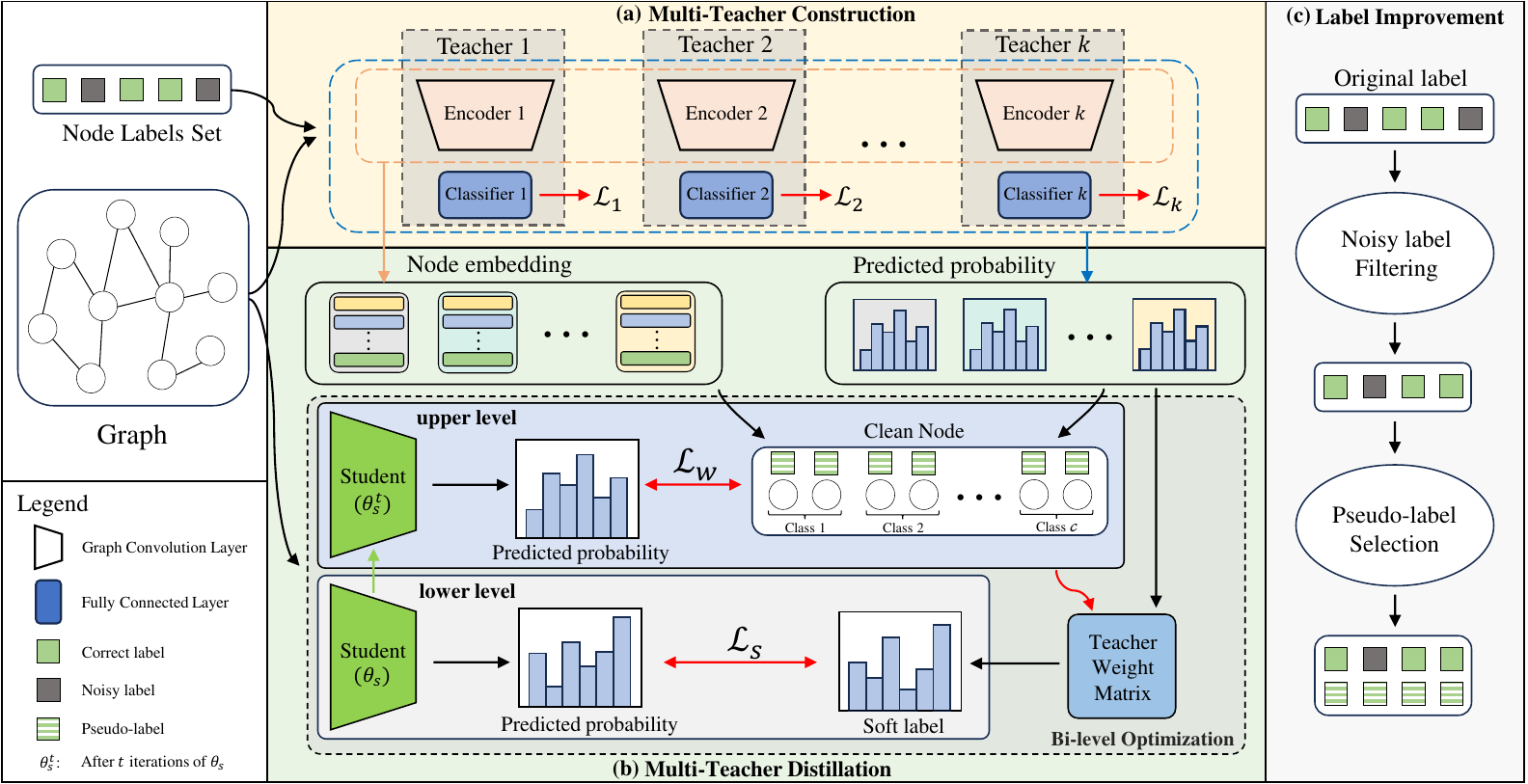}
    \caption{The framework of the proposed BO-NNC, consisting of three modules, \ie \textbf{Multi-teacher Construction}, \textbf{Multi-teacher Distillation}, and \textbf{Label Improvement}. Specifically, \textbf{Multi-teacher Construction}  employs multiple self-supervised  learning methods to obtain diverse teacher models. \textbf{Multi-teacher Distillation} transfers the knowledge from the teacher models to the student model through multi-teacher distillation based on bi-level optimization. Specifically, the lower level makes the student model learn the knowledge of the teacher models from the soft label matrix, which is  the Hadamard product between the teacher weight matrix and $k$ prediction probability matrices produced by teacher models. The upper level updates the teacher weight matrix based on the training progress of the student model. \textbf{Label Improvement} uses both the student model and the teacher models to first detect noisy labels and then select pseudo-labels.}
    \label{fig:structure}
\end{figure*}

Addressing the above issues of multi-teacher distillation methods is challenging. First, since the labelled data is sparse and with noisy, it is difficult to train diverse teacher models by existing methods (\eg randomly sampling training subsets from the training set). Second, previous methods usually require a large number of correct labels to learn the weights of the teacher models, which is not practical for NNC tasks. For example, MTS-GNN sets the weights of teacher models as hyper-parameters, resulting in expensive time cost and difficult to reasonably set the weights. 

In this paper, we propose a new multi-teacher distillation method based on bi-level optimization (namely BO-NNC, shown in Figure \ref{fig:structure}) to address the above issues. To do this, we first employ diverse self-supervised graph methods (\eg \cite{velivckovic2018deep,zhu2021graph,mo2022simple}) to construct multiple teacher models, solving the first issue of previous methods, and then construct a teacher weight matrix to integrate the predictions of the teacher models as the soft label matrix, which is regarded as the ground truth of the student model. Moreover, we propose a new bi-level optimization strategy to iteratively train the teacher weight matrix and the student model. This allows the teacher weight matrix to be dynamically adjusted based on the training progress of the student model without requiring a large number of correct labels to learn the weights of the teacher models. In particular, the proposed bi-level optimization algorithm helps to learn the complementary information among teacher models. After the bi-level optimization, we further design a label improvement module to improve the label quality of the dataset, which further enhances the learning of multiple teacher models and the student model.

Compared with previous NNC methods, the main contributions of our proposed method are summarized as follows:

\begin{itemize}[leftmargin=*]

\item[$\bullet$] We propose a new method to deal with noisy node classification. Specifically, our proposed method solves the limitations of previous methods by multi-teacher distillations, and designs a label improvement module to gradually improve the quality of labels.

\item[$\bullet$] We design a new distillation method based on bi-level optimization, which automatically adjust the teacher weight matrix, thus more fully exploring the complementary information among multiple teacher models.

\item[$\bullet$] We conduct extensive experiments on five real datasets. Experimental results demonstrate that our proposed method achieves the best performance, compared to existing SOTA methods at different levels of noisy data.
\end{itemize}

\section{Methodology}
Denoting $\mathcal{G}=(\mathcal{V}, \mathcal{E})$ as a graph, where $\mathcal{V}=\left\{v_{1}, \cdots, v_{n}\right\}$ represents the node set with $n$ nodes and $\mathcal{E} \subseteq \mathcal{V} \times \mathcal{V}$ represents the edge set, we denote $\mathbf{X} = \left\{\mathbf{x}_{1}, \cdots, \mathbf{x}_{n}\right\} \in \mathbb{R}^{n \times d}$ and $\mathbf{A} \in \mathbb{R}^{n \times n}$, respectively, as the feature matrix of all nodes and the adjacency matrix of the graph $\mathcal{G}$, where $\mathbf{x}_{i} \in \mathbb{R}^{d}$ is the $d$-dimensional feature vector of the $i$-th node. The  labels in the datasets are denoted as $\mathbf{Y} = \{\mathbf{y}_{1}, \cdots, \mathbf{y}_{b}\}$, where $b$ is the number of labelled nodes, $\mathbf{y}_{i} \in \mathbb{R}^{c}$ is one-hot encoding of the $i$-th label and $c$ is the number of classes.

\subsection{Multi-teacher Construction}
Existing studies \cite{wu2022automated} demonstrate that different representation learning methods usually output diverse node representations. Motivated by such observation, we  employ existing graph representation learning algorithms to construct multiple teacher models.

Specifically, we first train $k$ encoders by $k$ different unsupervised graph representation learning methods and then use them to obtain node embeddings:
\begin{eqnarray}
\mathbf{H}_{i} = \mathbf{F}_{i}(\mathbf{X}, \mathbf{A}),
\label{eq1}
\end{eqnarray}
where $i\in [1,..., k]$ and $\mathbf{F}_{i}$ denotes the $i$-th encoder. We then use individual embedding as inputs to train $k$ classifiers by the loss function as follows: 
\begin{eqnarray}
\mathcal{L}_{i} = -\sum_{j \in \mathcal{S}} \mathbf{y}_{j} \log \mathbf{p}_{j}^{i},
\label{eq2}
\end{eqnarray}
where $\mathcal{S}$ denotes the labelled training set, $\mathbf{p}_{j}^{i} \in \mathbb{R}^{c}$ denotes the prediction of the $i$-th classifier for node $v_j$, and $\mathbf{y}_{j}$ denotes the label of node $v_j$. In the end, we obtain $k$ teacher models consisting of $k$ encoders and $k$ classifiers to achieve model diversity. 

The literature (\eg \cite{hinton2015distilling,kim2021self}) demonstrate that the knowledge in the teacher model can be transferred to the student model by making the output of the student model mimic the output of the teacher model. 
Motivated by this, we consider using the predictions of the teacher models as soft labels to train the student model. This may enable the student model to explore the complementary information among teacher models, and thus obtaining better generalization performance. 
In this way, our method may successfully address the issue of the excessive dependence on the performance of a single model in previous methods.

\subsection{Multi-teacher Distillation} \label{Mtd}

After constructing multiple teacher models on labelled nodes, in this section, diverse knowledge will be transferred to the student model.
To do this, we first employ multiple teachers to generate the node embeddings $\mathbf{H} = \{\mathbf{H}_{1}, ..., \mathbf{H}_{k}\}$ and prediction probability matrices $\mathbf{P} = \{\mathbf{P}_{1}, ..., \mathbf{P}_{k}\}$ for all nodes (including labelled nodes and unlabelled nodes), which are then used to guide the student model.

Second, the student model are generated by randomly initialized parameters on all nodes. That is, the node embedding of all nodes $\mathbf{Z}^{(i)}$ in the $i$-th layer of the student model and the probability matrix $\mathbf{P}^S$ of the student model are obtained by:
\begin{equation}
\begin{cases}{}
\mathbf{Z}^{(i)} = \sigma(\mathbf{\hat{A}} \mathbf{Z}^{(i-1)} \mathbf{W}^{(i-1)}), \\
\mathbf{P}^S = \operatorname{softmax}(\sigma(\mathbf{\hat{A}} \mathbf{Z}^{(i)} \mathbf{W}^{(i)})),
\end{cases}
\label{eq3}
\end{equation}
where $\mathbf{W}^{(i)}$ is the trainable parameters of the $i$-th layer and $\sigma$ is the activation function.

After that, we fuse $k$ prediction probability matrices to a probability matrix, \ie soft label matrix, which is the ground truth of the student model.
The simple strategy for mapping $k$ prediction probability matrices to the soft label matrix is the mean method. However, such a method equivalently treats every teacher model.
A good alternative is co-attention method \cite{gao2021distilling}. Specifically, we learn the weight matrix $\mathcal{W} \in \mathbb{R}^{k \times c}$ ($c$ denotes the number of classes) to map $k$ prediction probability matrices to the soft label matrix. However, previous co-attention methods usually require a large number of correct labels to learn the weights of the teacher models, which is not feasible in NNC tasks.

To address the above issue, we investigate clean nodes to effectively learn the weight matrix, resulting in a bi-level optimization. Specifically, after $t$ iterations of the student model, we calculate the loss among clean nodes to update the teacher weight matrix by the back-propagation (Details in Section \ref{blo}). As a result,  information obtained from the student model can be used to improve  the teacher weight matrix. The updated teacher weight matrix guarantees the effectiveness of soft labels  even with low-quality teacher models due to their limited number of labels. 


In our method, clean nodes come from unlabelled nodes and labelled nodes. 
We select clean nodes from unlabelled nodes by first considering the prediction results of all teacher models and then filtering the selected clean nodes by node embedding similarity. 
Specifically, we first chose unlabelled nodes with consistent prediction among all teacher models to form the candidate set $Can$:
\begin{eqnarray}
Can = \{ v_i \mid \forall j \in (1,k-1), \, \arg\max\mathbf{p}_{i}^{j} = \arg\max\mathbf{p}_{i}^{k} \},
\label{eq4}
\end{eqnarray}
where $\mathbf{p}_{i}^{j}$ denotes the prediction of the $j$-th teacher model for node $v_i$.
We then filter clean nodes in $Can$ based on the node embedding similarity between every node in $Can$ and its $\beta_1$ ($\beta_1$ is a hyper-parameter) closest nodes in the same class, \ie
\begin{eqnarray}
\xi_i = \sum_{j=1}^{k}\sum_{v_u \in C_i} d(h^j_i, h^j_u),
\label{eq5}
\end{eqnarray}
where $C_i$ denotes the set consisting of $\beta_1$ closest nodes of the same class to node $v_i$, $h^j_i$ denotes the embedding vector of the $j$-th teacher for node $v_i$, and $d(\cdot, \cdot)$ denotes Euclidean distance. Finally, we select $\beta_2$ ($\beta_2$ is a hyper-parameter) nodes with the smallest $\xi$ (\ie with the highest confidence) in each class as the first part of clean nodes.

In order to increase the number of clean nodes, we consider selecting the second part of clean nodes from labelled nodes, based on the loss values of the teacher models. Specifically, we first use the teacher models to calculate loss values for all labelled nodes:
\begin{eqnarray}
{l}_{i}^{T} = -\sum_{j=1}^{k} \mathbf{y}_{i} \log \mathbf{p}_{i}^{j},
\label{eq6}
\end{eqnarray}
where ${l}_{i}^T$ denotes the loss value of node $v_i$. After that, we select $\alpha\%$ ($\alpha$ is a hyper-parameter) nodes with the lowest loss values (\ie with the highest confidence) in each class to form the second part of clean nodes. 

Finally, we combine two parts of clean nodes (\ie with the highest confidence out of all nodes) to obtain the clean node set $Cle$, which is used to calculate the loss of the upper level, in the bi-level optimization. Moreover, the loss in the upper level is then used to update the teacher weight matrix. In the lower level, the teacher weight matrix is used to generate the soft label matrix, which is further used to update the parameters of the student model. In particular, the bi-level optimization strategy makes the teacher weight matrix  be dynamically adjusted based on the feedback from the student model. This allows the updated teacher weight matrix  to effectively fuse multiple prediction matrices, \ie extracting their common information across all models and complementary information in individual models. 

\subsection{Label Improvement}
After conducting both multi-teacher construction and multi-teacher distillation, we design a label improvement module to gradually improve the label quality of the dataset during the training process. This module involves two phases, \ie noisy label filtering and pseudo-label selection.

\subsubsection{\textbf{Noisy label Filtering}}
Noisy label filtering is designed to filter a portion of noisy labels from the labelled data, thereby reducing incorrect supervision information in the dataset. Existing literature \cite{gui2021towards} shows that the node with higher loss has higher probability to be a noisy node than the node with lower loss. Moreover, the loss in one class is different from that in other classes. Hence, we calculate the loss in the $j$-th class between the node labels and the predictions of the student model by:
\begin{eqnarray}
L_j = \{ {l}_{i}^{S} \mid \mathbf{y}_i = j\}, ~\text{where}~ {l}_{i}^{S} = -\mathbf{y}_{i} \log \mathbf{p}_{i}^{S},
\label{eq18}
\end{eqnarray}
\eat{
Furthermore, existing studies have shown that there are differences in loss values between classes. Therefore, we consider filtering noisy labels separately for each class. Specifically, we separately take the loss values of labelled nodes of each class to compose a set:
\begin{eqnarray}
L_j = \{ {l}_{i}^{S} \mid y_i = j\},
\label{eq19}
\end{eqnarray}
}
where $j \in (1,c)$ denotes the $j$-th class. 
We then set a threshold $\boldsymbol{\varphi}$ for each class to select noisy nodes:
\begin{eqnarray}
\boldsymbol{\varphi}_{i} = sort(L_i)_{\lceil {\delta}_i r \rceil},
\label{eq20}
\end{eqnarray}
where ${\delta}_i$ denotes the number of labelled nodes of class $i$, $r$ is a hyper-parameter that denotes the percentage of selected noisy nodes in the total labelled nodes.
Finally, nodes with loss values larger than threshold in each class are regarded as noisy nodes and they are regarded as unlabelled nodes in the following part of the label improvement module.

\subsubsection{\textbf{Pseudo-label Selection}}
After removing noisy labels, pseudo-label selection investigates to assign pseudo-labels to unlabelled nodes, aim at increasing supervision information. We follow the literature \cite{li2018deeper} to separately select pseudo-labels based on the prediction confidence of both the student model and the teacher models.
Specifically, we first use the student model to predict all unlabelled nodes and then record the confidence of the student model by:
\begin{equation}
\delta^{S}_{i} = \max \mathbf{p}_{i}^{S},
\label{eq21}
\end{equation}
where $\delta^{S}_{i}$ denotes the prediction confidence of the student model for node $v_i$. Given the prediction confidence, we select the top $\rho$ ($\rho$ is a hyper-parameter) nodes with the highest $\delta^{S}$ in each class as the first part of pseudo-labelled nodes.

We then sum the predictions of the teacher models as teacher predictions and calculate the confidence of the teacher models by:
\begin{equation}
\delta^{T}_{i} = \max (\sum_{j=1}^k \mathbf{p}^j_i),
\label{eq22}
\end{equation}
where $\delta^{T}_{i}$ denotes the prediction confidence of the teacher models for node $v_i$. 
Similarly, we select the top $\rho$ nodes with the highest $\delta^{T}$ in each class as the second part of pseudo-labelled nodes. 

Finally, we assign  all  pseudo-labelled nodes with pseudo-labels and further add them to the labelled set, which will be used to adaptively update both the multi-teacher construction module and the multi-teacher distillation module, whose robustness are improved on the data with corrected labels and pseudo-labels with high quality. 


\section{Bi-level Optimization} \label{blo}
\subsection{Optimization Process}
Previous works often separately the teacher weight matrix and the student model in Section \ref{Mtd}. This  ignores the potentiality of the student model to provide information for improving the teacher weight matrix.
In this paper, we propose to dynamically adjust the teacher weight matrix based on the training progress of the student model, aiming at reducing the dependence of the learning the teacher weight matrix on correct labels. As a result, the optimization of both the teacher weight matrix and the student model results in a bi-level optimization problem,  \cite{liu2021investigating,chen2022gradient}.
Considering that the bi-level optimization usually includes
two levels of optimization tasks, where the solution of the upper optimization task depends on the results of the lower optimization task.
To address the above issue, in this paper, we treat the optimization of the teacher weight matrix as the upper level task and the training of the student model as the lower level task, to have the following objective function:

\begin{eqnarray}
\underbrace{\underset{\theta_{w}}{\arg\min}\ \mathbf{\Gamma}_{\scriptscriptstyle \text{up}} (\theta_{w} , \theta_{s}^{t})}_{\text{upper level task}} ~~~s.t.~~~ \theta_{s}^{t} = \underbrace{\underset{\theta_{s}}{\arg\min}\ \mathbf{\Gamma}_{\scriptscriptstyle \text{low}} (\theta_{w} , \theta_{s})}_{\text{lower level task}},
\label{eq7}
\end{eqnarray}
where $\mathbf{\Gamma}_{\scriptscriptstyle \text{up}}$ and $\mathbf{\Gamma}_{\scriptscriptstyle \text{low}}$, respectively, denote the loss functions for  the teacher weight matrix and  the student model. $\theta_{w}$ and $\theta_{s}$, respectively,  denote the trainable parameters of the teacher weight matrix $\mathcal{W}$ and the student model. $\theta_{s}^{t}$ denotes the parameters of the student model in the $t$-th iteration. 

In the lower-level optimization, we train the student model using soft labels. Specifically, we first construct a teacher weight matrix $\mathcal{W} \in \mathbb{R}^{k \times c}$ to fuse the predictions of  $k$ teacher models as a soft label matrix, \ie
\begin{eqnarray}
\widetilde{\mathbf{y}}_{i} = \operatorname{softmax}(\sum_{j=1}^{k} \mathcal{W}_j \odot \mathbf{p}_{i}^{j}),
\label{eq8}
\end{eqnarray}
where ``$\odot$" denotes the Hadamard product. We then calculate the loss of the lower level optimization by:
\begin{eqnarray}
\mathbf{\Gamma}_{\scriptscriptstyle \text{low}} (\theta_{w} , \theta_{s}) = \mathcal{L}_{s}, ~\text{where}~ \mathcal{L}_{s} = -\sum_{i \in \mathcal{T}} \widetilde{\mathbf{y}}_{i}\log \mathbf{p}_{i}^{S},
\label{eq9}
\end{eqnarray}
where $\mathcal{T}$ denotes the training set, and $\mathbf{p}_{i}^{S}$ denote the prediction of the student model for node $v_i$.

After calculating $\mathbf{\Gamma}_{\scriptscriptstyle \text{low}}$, we update the parameters $\theta_{s}$ of the student model by the gradient descent:
\begin{equation}
\theta_{s}^{\mu} = \theta_{s}^{\mu-1} - \eta_{\mu} \scalebox{1.3}{d}_{{\Gamma}_{\scriptscriptstyle \text{low}}}(\theta_{s}^{\mu-1}; \theta_{w}),
\label{eq10}
\end{equation}
where $\eta_{\mu}$ denotes the learning rate and $\scalebox{1.3}{d}_{{\Gamma}_{\scriptscriptstyle \text{low}}}(\theta_{s}^{\mu-1}; \theta_{w})$ denotes the derivative of $\mathcal{L}_{s}$ with respect to $\theta_{s}^{\mu-1}$. As a result, in the lower level optimization, we transfer the knowledge from the teacher models to the student model by having the prediction of the student model mimic the soft labels. 

After  the lower level optimization, we first pass the result $\theta_{s}^{t}$ of the lower level optimization to the upper level optimization, and then conduct the upper-level optimization. To do this, we update the teacher weight matrix using both the student model (after $t$ iterations) and clean nodes. Specifically, the loss function of the upper level optimization is designed as the cross-entropy loss between the prediction of  the clean nodes in the student model and the pseudo-labels of  clean nodes:
\begin{eqnarray}
\mathbf{\Gamma}_{\scriptscriptstyle \text{up}} (\theta_{w} , \theta_{s}^{t}) = \mathcal{L}_{w}, ~\text{where}~ \mathcal{L}_{w} = -\sum_{i \in Cle} \hat{\mathbf{y}}_{i} \log \mathbf{p}^{s_{t}}_{i},
\label{eq11}
\end{eqnarray}
where $Cle$ denotes the clean node set, $\hat{\mathbf{y}}_{i}$ denotes the pseudo-label of $v_i$, and $\mathbf{p}^{s_{t}}_{i}$ denotes the prediction of the student model (after $t$ iterations) for node $v_i$.

Obviously, $\theta_{w}$ is not directly involved in the calculation of $\mathcal{L}_{w}$, thus we cannot follow the ordinary approach to directly calculate the gradient of $\mathcal{L}_{w}$ with respect to $\theta_{w}$ and update $\theta_{w}$. To address this issue, we follow the literature \cite{liu2021investigating,liu2021value} to perform  derivations based on  specific situation.

Specifically, we denote the outer optimization objective as $\underset{\theta_{w}}{\min}\ \boldsymbol{\varphi} (\theta_{w}) := \mathbf{\Gamma}_{\scriptscriptstyle \text{up}} (\theta_{w} , \theta_{s}^{t})$, where  $\theta_{s}^{t}$ is a variable dependent on $\theta_{w}$, so we have:
\begin{eqnarray}
\frac{\partial {\boldsymbol{\varphi} (\theta_{w})}}{\partial {\theta_{w}}} = \underbrace{\frac{\partial {\mathbf{\Gamma}_{\scriptscriptstyle \text{up}} (\theta_{w} , \theta_{s}^{t})}}{\partial {\theta_{w}}}}_{\text{direct grad}} + \underbrace{\frac{\partial {\mathbf{\Gamma}_{\scriptscriptstyle \text{up}} (\theta_{w} , \theta_{s}^{t})}}{\partial {\theta_{s}^{t}}} (\frac{\partial {\theta_{s}^{t}}}{\partial {\theta_{w}}})}_{\text{indirect grad}}.
\label{eq12}
\end{eqnarray}

For the  Eq.\ref{eq12}, on the one hand,  $\theta_{w}$ is not directly involved in the calculation of $\mathcal{L}_{w}$, thus the ``direct grad" is zero. On the other hand, the ``indirect grad" consists of two parts, \ie $\frac{\partial {\mathbf{\Gamma}_{\scriptscriptstyle \text{up}} (\theta_{w} , \theta_{s}^{t})}}{\partial {\theta_{s}^{t}}}$ and $\frac{\partial {\theta_{s}^{t}}}{\partial {\theta_{w}}}$. Since the former can be directly calculated  by the derivation of $\mathcal{L}_{w}$ with respect to $\theta_{s}^{t}$, the main challenging lies in the calculation both $\frac{\partial {\theta_{s}^{t}}}{\partial {\theta_{w}}}$ and $\frac{\partial {\boldsymbol{\varphi} (\theta_{w})}}{\partial {\theta_{w}}}$.

In this paper, we first dynamically unfold $\theta_{s}^{t}$ based on Eq.\ref{eq10} to calculate $\frac{\partial {\theta_{s}^{t}}}{\partial {\theta_{w}}}$ by:
\begin{gather}
\label{eq13}
\theta_{s}^{t} = \boldsymbol{\psi}_{t}(\theta_{s}^{t-1}; \theta_{w}), ~~
\text{where}~~
\\ \nonumber
\boldsymbol{\psi}_{i}(\theta_{s}^{i-1}; \theta_{w}) = \theta_{s}^{i-1} - \eta_{i} \scalebox{1.3}{d}_{\mathbf{\Gamma}_{\scriptscriptstyle \text{low}}}(\theta_{s}^{i-1}; \theta_{w}),~i=1, \cdots, t.
\end{gather}

Combining $\frac{\partial {\theta_{s}^{t}}}{\partial {\theta_{w}}}$ and Eq.\ref{eq13}, we have:
\begin{eqnarray}
\frac{\partial {\theta_{s}^{t}}}{\partial {\theta_{w}}} = \frac{\partial {\boldsymbol{\psi}_{t}(\theta_{s}^{t-1}; \theta_{w})}}{\partial {\theta_{s}^{t-1}}} \frac{\partial {\theta_{s}^{t-1}}}{\partial {\theta_{w}}} + \frac{\partial {\boldsymbol{\psi}_{t}(\theta_{s}^{t-1}; \theta_{w})}}{\partial {\theta_{w}}}.
\label{eq14}
\end{eqnarray}

To simplify the notation, we denote:
\begin{equation}
\mathbf{Z}_t = \frac{\partial {\theta_{s}^{t}}}{\partial {\theta_{w}}},\ 
\mathbf{A}_t = \frac{\partial {\boldsymbol{\psi}_{t}(\theta_{s}^{t-1}; \theta_{w})}}{\partial {\theta_{s}^{t-1}}},\ 
\mathbf{B}_t = \frac{\partial {\boldsymbol{\psi}_{t}(\theta_{s}^{t-1}; \theta_{w})}}{\partial {\theta_{w}}}.
\label{eq15}
\end{equation}

We then rewrite Eq.\ref{eq14} to:
\begin{eqnarray}
\frac{\partial {\theta_{s}^{t}}}{\partial {\theta_{w}}} = \mathbf{Z}_t = \sum \limits_{i=0}^{t}{(\prod \limits_{j = 0}^{t} \mathbf{A}_j)\mathbf{B}_i},
\label{eq16}
\end{eqnarray}
where $\mathbf{A}_0 = \mathbf{B}_0 = 0$. $\mathbf{A}_i$ and $\mathbf{B}_i$ ($i \neq 0$), respectively, can be obtained by calculating the derivative of $\scalebox{1.3}{d}_{\mathbf{\Gamma}_{\scriptscriptstyle \text{low}}}(\theta_{s}^{i-1}; \theta_{w})$ with respect to $\theta_{s}^{i-1}$ and $\theta_{w}$.

Based on the above derivation, it is obviously that we can iteratively calculate $\frac{\partial {\theta_{s}^{t}}}{\partial {\theta_{w}}}$ during the optimization of the lower level optimization. Finally, we calculate $\frac{\partial {\boldsymbol{\varphi} (\theta_{w})}}{\partial {\theta_{w}}} = \frac{\partial {\mathbf{\Gamma}_{\scriptscriptstyle \text{up}} (\theta_{w} , \theta_{s}^{t})}}{\partial {\theta_{s}^{t}}} (\frac{\partial {\theta_{s}^{t}}}{\partial {\theta_{w}}})$ and update the parameters $\theta_{w}$ of the teacher weight matrix by the gradient descent method:
\begin{equation}
\theta_{w}^{\lambda} = \theta_{w}^{\lambda-1} - \eta_{\lambda} \scalebox{1.3}{d}_{\boldsymbol{\varphi} (\theta_{w})}(\theta_{s}^{t}; \theta_{w}^{\lambda-1}),
\label{eq17}
\end{equation}
where $\eta_{\lambda}$ denotes the learning rate and $\scalebox{1.3}{d}_{\boldsymbol{\varphi} (\theta_{w})}(\theta_{s}^{t}; \theta_{w}^{\lambda-1})$ denotes the derivative of $\mathcal{L}_{w}$ with respect to $\theta_{w}^{\lambda-1}$. 

In the upper level optimization, we use the feedback  from the lower level optimization (\ie the student model) to update the teacher weight matrix, and thus rationally adjust the weights of the teacher models. After updating the teacher weight matrix,  we map the predictions of  $k$ teacher models to soft labels through Eq.\ref{eq8} and then train the student models again (\ie the lower level optimization). By iteratively updating the upper level optimization and lower level optimization, we fully transfer knowledge from the teacher models to the student model. 

Finally, through bi-level optimization strategy, we successfully addressed the issue that previous methods require a large number of correct labels to learn the weights of the teacher models (Validated in Section \ref{Bi-level optimization is useful}).

\subsection{Complexity Analysis}
\subsubsection{\textbf{Time Complexity}}
The bi-level optimization needs to first construct upper and lower level optimization tasks and and then iteratively optimize these two tasks. 
Specifically, in the forward propagation, our method needs the time cost of O($n^2d$), where $n$ and $d$, respectively, are the number of nodes and dimensions. In the  back propagation, our bi-level optimization requires to calculate the gradient $\frac{\partial {\theta_{s}^{t}}}{\partial {\theta_{w}}}$ and its complexity is O($s+w$) \cite{franceschi2017forward}, where $w$ is the parameter number of the teacher weight matrix, usually $s >> w$.
As a result, the time complexity of our bi-level optimization method is O($n^2d+s+w$). In contrast, traditional optimization methods need O($n^2d+s$). Obviously, they are quadratic to the node number.


\subsubsection{\textbf{Space Complexity}}

The bi-level optimization strategy calculates the gradients of the loss for the parameters of the student model to have the space complexity of 
O($(n+d+c)n+s$), where the space complexity  of the feature matrix, the adjacency matrix,  the prediction matrix,  the model parameter matrix, and the parameter gradient matrix, respectively, are with O($nd$), O($n^2$),  O($nc$), O($s$), and O($s$), where $c$ denotes the output dimension of the model. 
The bi-level optimization strategy need to store $\frac{\partial {\theta_{s}^{t}}}{\partial {\theta_{w}}}$, $\frac{\partial {\boldsymbol{\psi}_{t}(\theta_{s}^{t-1}; \theta_{w})}}{\partial {\theta_{s}^{t-1}}}$ and $\frac{\partial {\boldsymbol{\psi}_{t}(\theta_{s}^{t-1}; \theta_{w})}}{\partial {\theta_{w}}}$ in the lower level optimization task as well. Hence, the  space complexity in our bi-level optimization strategy is O($(n+d+c)n+s+w$), which is quadratic to the node number. 
In contrast, the space complexity of the traditional optimization strategy (\eg the first-order gradient descent methods) is of O($(n+d+c)n+s$).

\section{Experiments}

\renewcommand\arraystretch{1}
\tabcolsep = 0.3cm
\begin{table*}[!t]
  \centering
  \normalsize
  \caption{Classification accuracy (average accuracy (\%) and standard deviation) of all methods with different noise rates on all datasets. Note that, the bold number represents the best results in the whole column.}
  {
    \begin{tabular}{c|c|c c c|c c}
    \cline{1-7}
    \multirow{2}{*}{Datasets} & \multirow{2}{*}{Methods} & \multicolumn{3}{c|}{Uniform} & \multicolumn{2}{c}{Pair} \\
\cline{3-7}          &       & 20\%   & 40\%   & 60\%   & 20\%   & 40\% \\
    \cline{1-7}
    \rule{0pt}{10pt}
    \multirow{7}{*}{Cora} & GCN   & 74.74 (0.6) & 69.40 (1.6) & 41.76 (1.9) & 72.58 (2.1) & 51.82 (0.4) \\
          & DGI   & 81.56 (0.2) & 79.92 (0.4) & 48.94 (0.3) & 81.16 (0.3) & 61.60 (0.5) \\
          & GREET & 80.88 (0.8) & 78.22 (1.3) & 44.70 (2.8) & 77.96 (1.7) & 56.42 (1.8) \\
          & JoCoR & 75.58 (0.8) & 71.14 (1.1) & 42.82 (1.7) & 74.46 (1.1) & 52.54 (1.2) \\
          & NRGNN & 79.14 (0.7) & 78.68 (1.0) & 52.02 (2.1) & 78.66 (1.8) & 56.30 (5.9) \\
          & MTS-GNN & 81.50 (0.8) & 79.26 (0.7) & 63.34 (3.1) & 80.68 (1.8) & 62.16 (2.2) \\
          & \textbf{Proposed} & {\textbf{82.64 (1.1)}} & {\textbf{82.20 (0.9)}} & {\textbf{73.16 (1.0)}} & {\textbf{81.64 (1.6)}} & {\textbf{73.46 (3.2)}} \\
    \cline{1-7}
    \rule{0pt}{10pt}
    \multirow{7}{*}{Citeseer} & GCN   & 63.18 (0.6) & 56.04 (2.0) & 37.74 (0.7) & 66.60 (0.9) & 40.68 (2.7) \\
          & DGI   & 69.24 (0.2) & 63.20 (0.1) & 42.14 (0.3) & 69.24 (0.3) & 45.14 (0.2) \\
          & GREET & 68.36 (0.9) & 62.46 (1.2) & 44.50 (1.7) & 70.74 (1.5) & 46.88 (1.0) \\
          & JoCoR & 69.06 (1.2) & 57.18 (1.4) & 38.64 (1.0) & 69.54 (0.6) & 46.82 (2.4) \\
          & NRGNN & 66.22 (1.2) & 59.66 (1.5) & 45.82 (3.2) & 67.86 (1.8) & 50.10 (2.1) \\
          & MTS-GNN & 71.94 (1.6) & 69.74 (1.1) & 53.64 (2.1) & 72.34 (1.1) & 59.84 (2.8) \\
        & \textbf{Proposed} & {\textbf{72.14 (0.7)}} & {\textbf{70.80 (0.8)}} & {\textbf{58.18 (3.9)}} & {\textbf{72.90 (0.5)}} & {\textbf{63.36 (3.9)}} \\
    \cline{1-7}
    \rule{0pt}{10pt}
    \multirow{7}{*}{DBLP} & GCN   & 80.32 (0.2) & 77.54 (1.1) & 65.05 (0.6) & 80.19 (0.5) & 61.56 (1.2) \\
          & DGI   & 79.86 (0.2) & 79.28 (0.2) & 67.59 (0.2) & 78.99 (0.1) & 70.29 (0.5) \\
          & GREET & 80.62 (0.3) & 79.83 (0.5) & 69.10 (1.0) & 81.16 (0.2) & 71.41 (1.8) \\
          & JoCoR & 80.77 (0.3) & 78.52 (0.9) & 67.32 (1.1) & 80.26 (0.3) & 64.71 (3.3) \\
          & NRGNN & 81.89 (0.4) & 82.20 (0.3) & 76.13 (1.6) & 82.01 (0.4) & 72.80 (1.2) \\
          & MTS-GNN & 82.76 (0.2) & 82.49 (0.3) & 78.82 (1.5) & 82.68 (2.4) & 70.35 (4.5) \\
        & \textbf{Proposed} & {\textbf{83.93 (0.2)}} & {\textbf{83.87 (0.2)}} & {\textbf{81.18 (0.3)}} & {\textbf{83.56 (1.7)}} & {\textbf{77.32 (0.8)}} \\
    \cline{1-7}
    \multirow{7}{*}{Photo} & GCN   & 88.15 (0.6) & 84.51 (0.7) & 73.27 (5.9) & 88.50 (0.8) & 66.20 (1.3) \\
          & DGI   & 87.83 (0.4) & 86.78 (0.5) & 78.61 (0.9) & 85.67 (0.1) & 70.71 (0.2) \\
          & GREET & 88.77 (0.4) & 86.39 (0.5) & 74.71 (1.2) & 88.06 (0.4) & 65.66 (0.7) \\
          & JoCoR & 88.34 (1.1) & 84.17 (0.8) & 79.22 (2.1) & 88.78 (0.5) & 70.02 (2.9) \\
          & NRGNN & 88.30 (0.5) & 86.17 (0.6) & 77.25 (3.5) & 88.61 (0.6) & 66.20 (2.4) \\
          & MTS-GNN & 91.01 (1.4) & 88.10 (1.2) & 85.12 (1.7) & {\textbf{91.74 (0.3)}} & 81.26 (4.6) \\
        & \textbf{Proposed} & {\textbf{91.75 (0.1)}} & {\textbf{90.06 (1.0)}} & {\textbf{87.01 (0.9)}} & 91.26 (0.3) & {\textbf{86.96 (2.3)}} \\
    \cline{1-7}
    \rule{0pt}{10pt}
    \multirow{7}{*}{Computers} & GCN & 86.66 (0.3) & 83.18 (0.5) & 78.07 (0.4) & 84.60 (0.1) & 72.91 (0.8) \\
          & DGI   & 83.12 (0.1) & 77.71 (0.1) & 75.89 (0.3) & 81.73 (0.2) & 77.71 (0.1) \\
          & GREET & 84.16 (0.2) & 80.75 (0.8) & 72.19 (1.2) & 83.05 (0.1) & 64.35 (0.9) \\
          & JoCoR & {\textbf{86.96 (0.4)}} & 83.46 (0.4) & 77.99 (0.5) & 84.92 (0.4) & 73.93 (0.9) \\
          & NRGNN & 86.51 (0.5) & 83.63 (0.5) & 77.37 (0.8) & 84.34 (0.5) & 74.22 (1.1) \\
          & MTS-GNN & 86.43 (0.2) & 84.29 (0.2) & 80.65 (0.1) & {\textbf{86.36 (0.4)}} & 76.55 (2.1) \\
        & \textbf{Proposed} & 85.81 (0.2) & {\textbf{84.89 (0.1)}} & {\textbf{80.85 (0.3)}} & 86.04 (0.2) & {\textbf{82.38 (1.2)}} \\
    \cline{1-7}
    \end{tabular}
    \label{Tabel1}
    }
\end{table*}%

\subsection{Experimental Settings}

\subsubsection{\textbf{Datasets}}
The benchmark datasets in our experiments include three citation datasets (\ie DBLP \cite{bojchevski2017deep}, Cora and Citeseer \cite{yang2016revisiting}), and two business datasets (\ie Computers and Photo \cite{shchur2018pitfalls}).

\subsubsection{\textbf{Comparison Methods}}
The comparison methods include two baseline methods (\ie GCN \cite{kipf2016semi} and GAT \cite{velivckovic2017graph}), four unsupervised graph representation learning method (\ie DGI \cite{velivckovic2018deep}, GCA \cite{zhu2021graph}, SUGRL \cite{mo2022simple}, and GREET \cite{liu2023beyond}), and three noisy label methods (\ie JoCoR \cite{wei2020combating}, NRGNN \cite{dai2021nrgnn}, and MTS-GNN \cite{liu2023multi}).

\renewcommand\arraystretch{1} 
\tabcolsep = 0.4cm 
\begin{table*}[!t]
  \centering 
  \normalsize
  \caption{Classification accuracy (average accuracy (\%) and standard deviation) of the proposed BO-NNC with different components on all datasets at the highest noise rates for the two noise types. Note that the bold number represents the best results in the whole column.}
  {
    \begin{tabular}{ccc|ccccc} 
    \cline{1-8}
    \multicolumn{8}{c}{Uniform (60\%)}\\
    \cline{1-8}
    C1 & C2 & C3 & Cora & Citeseer & DBLP & Photo & Computers\\
    \cline{1-8}
    
    \checkmark &            &            & 55.22 (1.0) & 44.74 (2.4) & 77.57 (0.8) & 83.74 (1.3) & 79.54 (0.4) \\
    
               &            & \checkmark & 72.50 (1.3) & 53.24 (3.0) & 80.54 (0.4) & 86.23 (1.1) & 80.45 (0.6) \\
    
    \checkmark & \checkmark &            & 58.56 (5.0) & 45.28 (2.1) & 76.67 (0.6) & 84.04 (1.0) & 79.30 (2.0) \\
    
    \checkmark &            & \checkmark & 71.26 (1.4) & 56.78 (3.5) & 81.25 (0.3) & 86.10 (0.7) & 80.36 (0.7) \\

    \checkmark & \checkmark & \checkmark & \textbf{73.16 (1.0)} & \textbf{58.18 (3.9)} & \textbf{81.95 (0.3)} & \textbf{87.01 (0.9)} & \textbf{80.85 (0.3)} \\
    \cline{1-8}
    \end{tabular}%
    \\
    \rule{0pt}{3pt}
    \\ 
    \begin{tabular}{ccc|ccccc} 
    \cline{1-8}
    \multicolumn{8}{c}{Pair (40\%)}\\
    \cline{1-8}
    C1 & C2 & C3 & Cora & Citeseer & DBLP & Photo & Computers\\
    \cline{1-8}
    \checkmark &            &            & 67.14 (3.1) & 51.56 (2.3) & 76.05 (1.0) & 78.70 (5.2) & 79.42 (1.1) \\
    
               &            & \checkmark & 70.52 (6.4) & \textbf{63.56 (2.0)} & 74.33 (0.7) & 81.08 (1.2) & 81.09 (0.6) \\
    
    \checkmark & \checkmark &            & 65.76 (5.3) & 55.62 (2.1) & 75.22 (1.8) & 79.39 (5.0) & 76.43 (2.3) \\
    
    \checkmark &            & \checkmark & 73.36 (3.5) & 62.08 (2.5) & 76.01 (1.0) & 84.95 (2.2) & 81.35 (1.3) \\

    \checkmark & \checkmark & \checkmark & \textbf{77.50 (2.9)} & 63.36 (3.9) & \textbf{77.32 (0.8)} & \textbf{86.96 (2.3)} & \textbf{82.38 (1.2)} \\
    \cline{1-8}
    \end{tabular}%
    \label{tab:ablation}%
    }
\end{table*}

\renewcommand\arraystretch{1} 
\tabcolsep = 0.4cm 
\begin{table*}[!t]
  \centering 
  \normalsize
  \caption{Classification accuracy (average accuracy (\%) and standard deviation) of different multi-teacher distillation methods at 60\% noise rates on all datasets. Note that, the bold number represents the best results in the whole column.}
  {
    \begin{tabular}{c|ccccc} 
    \cline{1-6}
    \multicolumn{6}{c}{Uniform (60\%)}\\
    \cline{1-6}
    & Cora & Citeseer & DBLP & Photo & Computers\\
    \cline{1-6}
    Mean & 71.26 (1.4) & 56.78 (3.5) & 81.25 (0.3) & 86.10 (0.7) & 80.36 (0.7) \\
    
    Co-attention & 71.82 (0.9) & 57.24 (3.2) & 81.35 (0.3) & 85.66 (1.7) & 80.78 (0.6) \\
    
    \textbf{Proposed} & \textbf{73.16 (1.0)} & \textbf{58.18 (3.9)} & \textbf{81.95 (0.3)} & \textbf{87.01 (0.9)} & \textbf{80.85 (0.3)} \\
    \cline{1-6}
    \end{tabular}%
        \\
    \rule{0pt}{3pt}
    \\ 
    \begin{tabular}{c|ccccc} 
    \cline{1-6}
    \multicolumn{6}{c}{Pair (40\%)}\\
    \cline{1-6}
    & Cora & Citeseer & DBLP & Photo & Computers\\
    \cline{1-6}
    Mean & 73.36 (3.5) & 62.08 (2.5) & 76.01 (1.0) & 84.95 (2.2) & 81.35 (1.3) \\
    
    Co-attention & 72.96 (0.4) & 62.10 (2.7) & 77.17 (0.9) & 85.08 (1.9) & 80.80 (1.6) \\
    
    \textbf{Proposed} & \textbf{77.50 (2.9)} & \textbf{63.36 (3.9)} & \textbf{77.32 (0.8)} & \textbf{86.96 (2.3)} & \textbf{82.38 (1.2)} \\
    \cline{1-6}
    \end{tabular}%
    \label{tab:Bo_useful}%
    }
\end{table*}%

\subsubsection{\textbf{Setting-up}}
We  follow the PyTorch Geometric library to split the datasets Cora and Citeseer,  and randomly sample 5\%, 10\%, and 60\% of the nodes, respectively,   as labelled training set, validation set, and test set for the datasets DBLP, Photo, and Computers. There is no overlap among the training set, the validation set and the test set.
We follow the literature \cite{han2018co,jiang2018mentornet} to add different rates of noisy into the real datasets. Specifically, given a noise rate $p$, we first generate noisy labels over all classes according to a noise transition matrix $\mathbf{Q} \in \mathbb{R}^{c \times c}$, where $q_{ij} = \mathbf{p} (\overline{y} = j|y = i)$ is the probability of clean label $y$ being flipped to the noisy label $\overline{y}$, and then corrupt the labels of label training set with two types of label noise:
\begin{itemize}[leftmargin=*]
\item[$\bullet$]\textbf{Uniform noise} The label has $p$ probability of being randomly flipped to other class. Furthermore, we set noise rates of 20\%, 40\%, and 60\% for the uniform noise.
\item[$\bullet$]\textbf{Pair noise} Labels are assumed to make mistakes only within the most similar pair classes. More specifically, label has  $p$ probability of being randomly flipped to their pair classes. Furthermore, we set noise rates of 20\%, 40\% for the pair noise.
\end{itemize}

To reduce randomness of experimental results, we conduct five experiments with different random seeds and report the average results and corresponding standard deviations. In our experiments, we train three encoders by employing GCA \cite{zhu2021graph}, DGI \cite{velivckovic2018deep} and SUGRL \cite{mo2022simple}, plus single layer MLP for every encoder to obtain three teacher classifiers. For a fair comparison, the methods, including JoCoR, NRGNN, MTS-GNN, and the student model of our method use a two-layer GCN as the backbone. In addition, we obtain the source code of all comparison methods from the authors and set the parameters of all comparison methods according to the original literature so that they output the best performance on all datasets.

\subsection{Result Analysis}

We compare our proposed BO-NNC with all comparison methods on five datsets in terms of node classification tasks with different noise rates and report the results in Table \ref{Tabel1}.

First, the proposed BO-NNC consistently achieves the best results on all datasets, followed by MTS-GNN, NRGNN, DGI, GREET, JoCoR, GAT, GCA, SUGRL and GCN. 
The reason is that our BO-NNC enables the student model to obtain the knowledge from multiple teacher models, thus solving the limitations of the single model to effectively deal with noisy labels.
Second, our BO-NNC outperforms two baseline methods (\ie GCN and GAT) and four unsupervised graph representation learning method (\ie GCA, DGI, SUGRL, and GREET) by a large margin, because  the proposed method provides more correct supervised information for the model training by noisy label filtering and pseudo-label selection. For example, the proposed BO-NNC averagely improves by 2.93\%, 4.99\%, and 13.44\% respectively, compared with the DGI that performs best in there six methods, in uniform noisy rates of 20\%, 40\%, and 60\% on all datasets. 
Third, our MTS-GNN achieves promising improvements, compared with the methods that deal with the noisy label problem, \ie JoCoR, NRGNN, and MTS-GNN. 
This indicates that our method fully leverages the complementary information among teacher models through bi-level optimization, and thus enabling the student model to achieve better generalization performance.

    
    
    
    

\begin{figure*}[t]
    \centering
    \subfigure[Cora]{
    \begin{minipage}[t]{0.15\linewidth}
    \centering
    \includegraphics[width=\linewidth, trim = 20 80 30 120, clip]{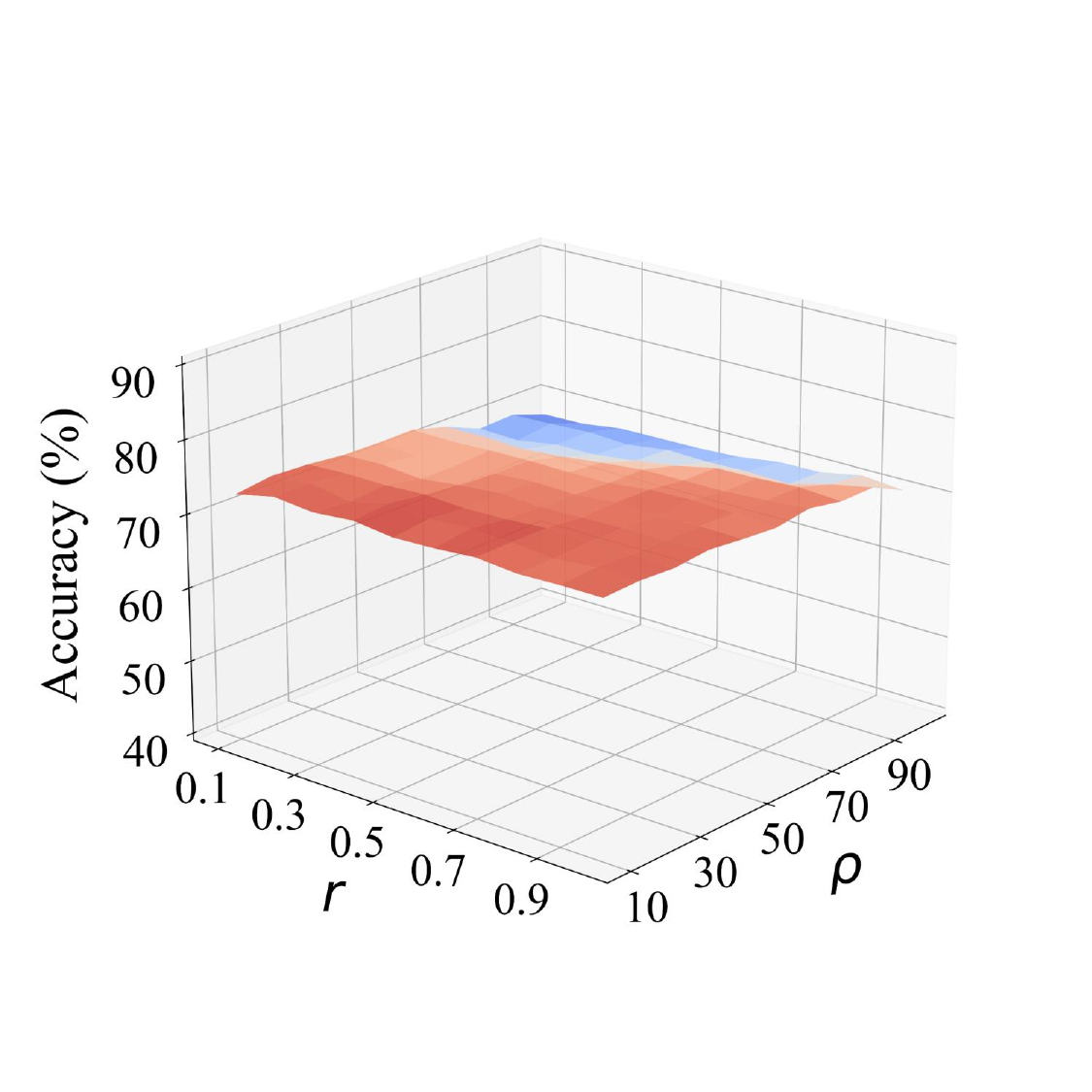}
    \end{minipage}%
    }%
    \hspace{0.01\textwidth}
    \subfigure[Citeseer]{
    \begin{minipage}[t]{0.15\linewidth}
    \centering
    \includegraphics[width=\linewidth, trim = 20 80 30 120, clip]{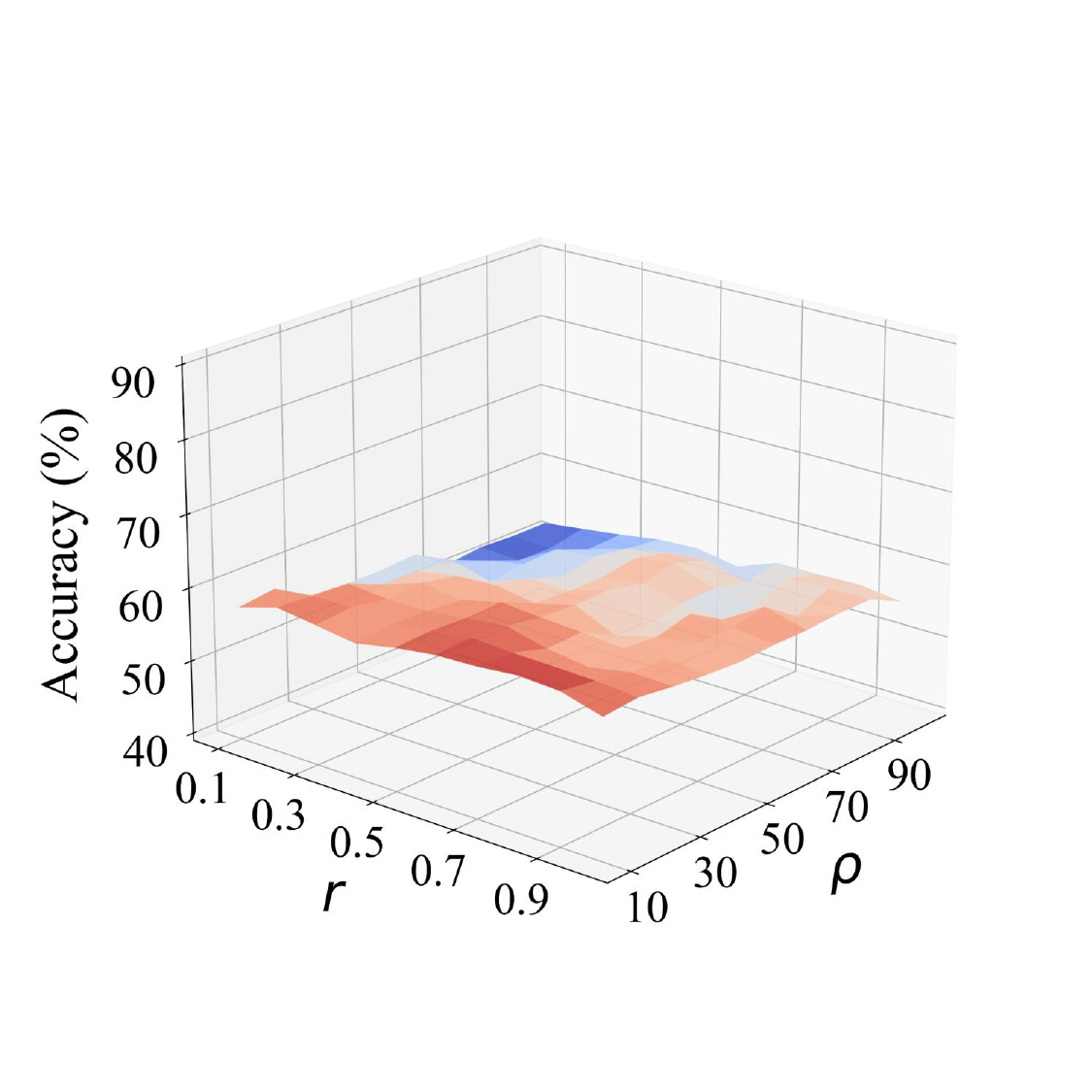}
    \end{minipage}%
    }%
    \hspace{0.01\textwidth}
    \subfigure[DBLP]{
    \begin{minipage}[t]{0.15\linewidth}
    \centering
    \includegraphics[width=\linewidth, trim = 20 80 30 120, clip]{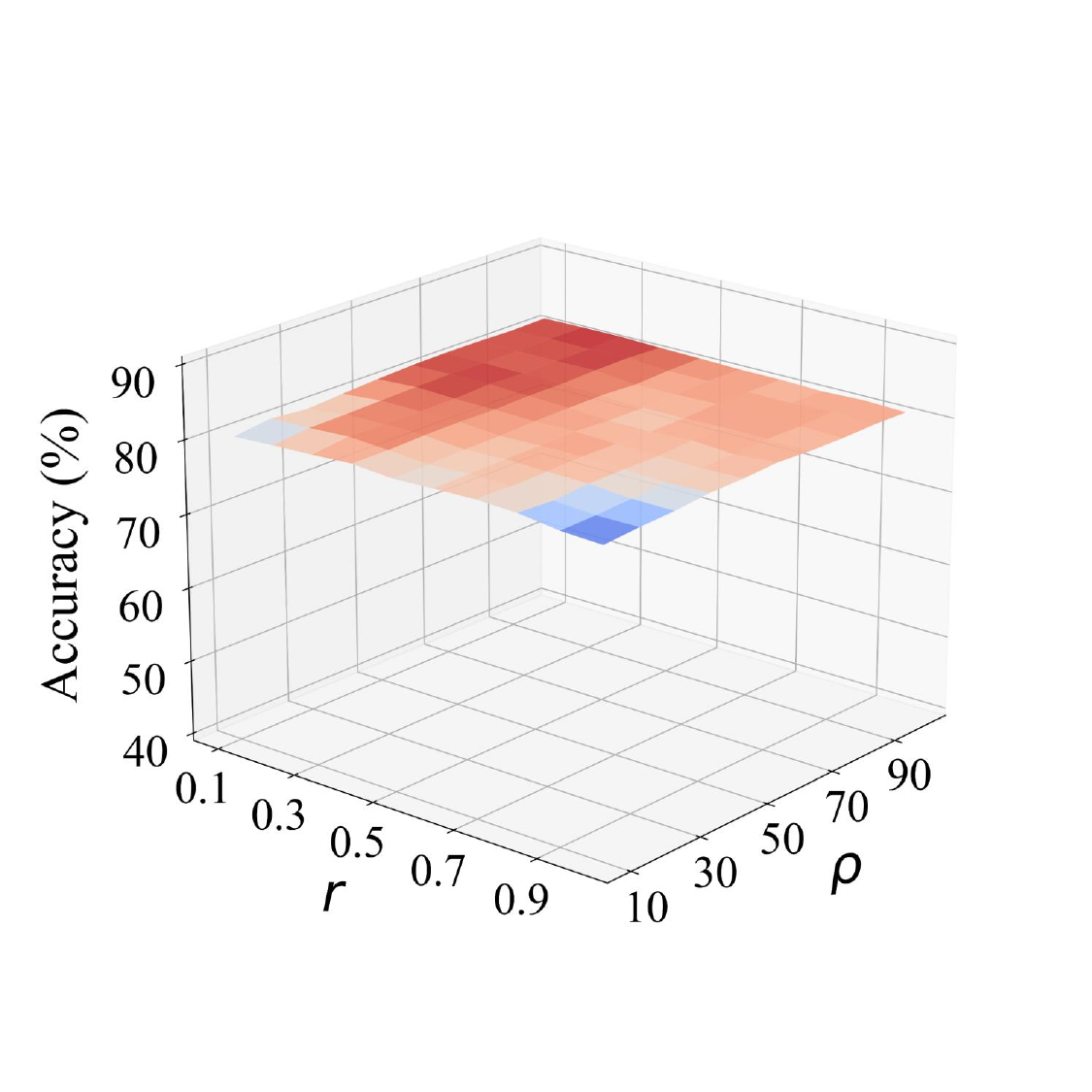}
    \end{minipage}%
    }%
    \hspace{0.01\textwidth}
    \subfigure[Photo]{
    \begin{minipage}[t]{0.15\linewidth}
    \centering
    \includegraphics[width=\linewidth, trim = 20 80 30 120, clip]{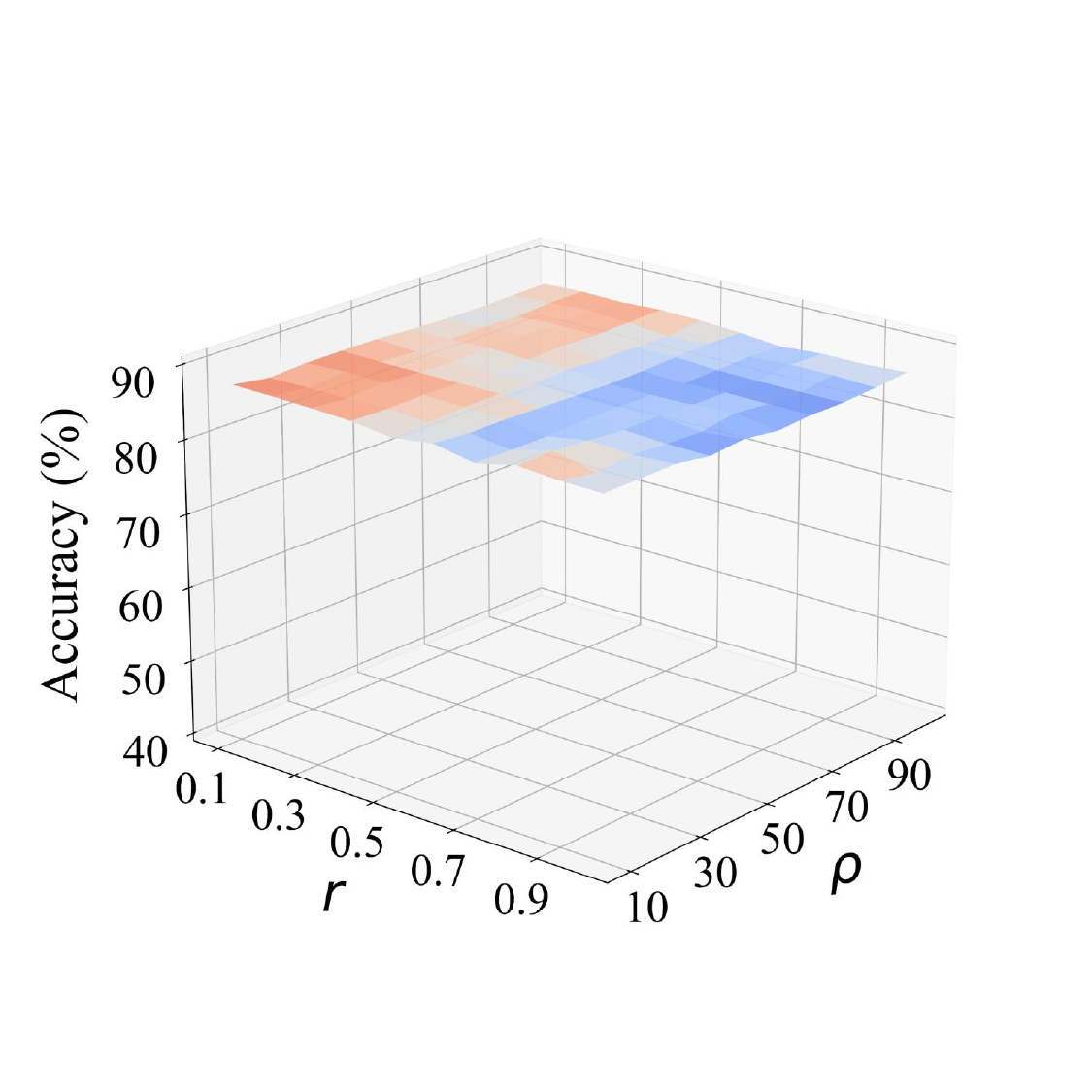}
    \end{minipage}%
    }%
    \hspace{0.01\textwidth}
    \subfigure[Computers]{
    \begin{minipage}[t]{0.15\linewidth}
    \centering
    \includegraphics[width=\linewidth, trim = 20 80 30 120, clip]{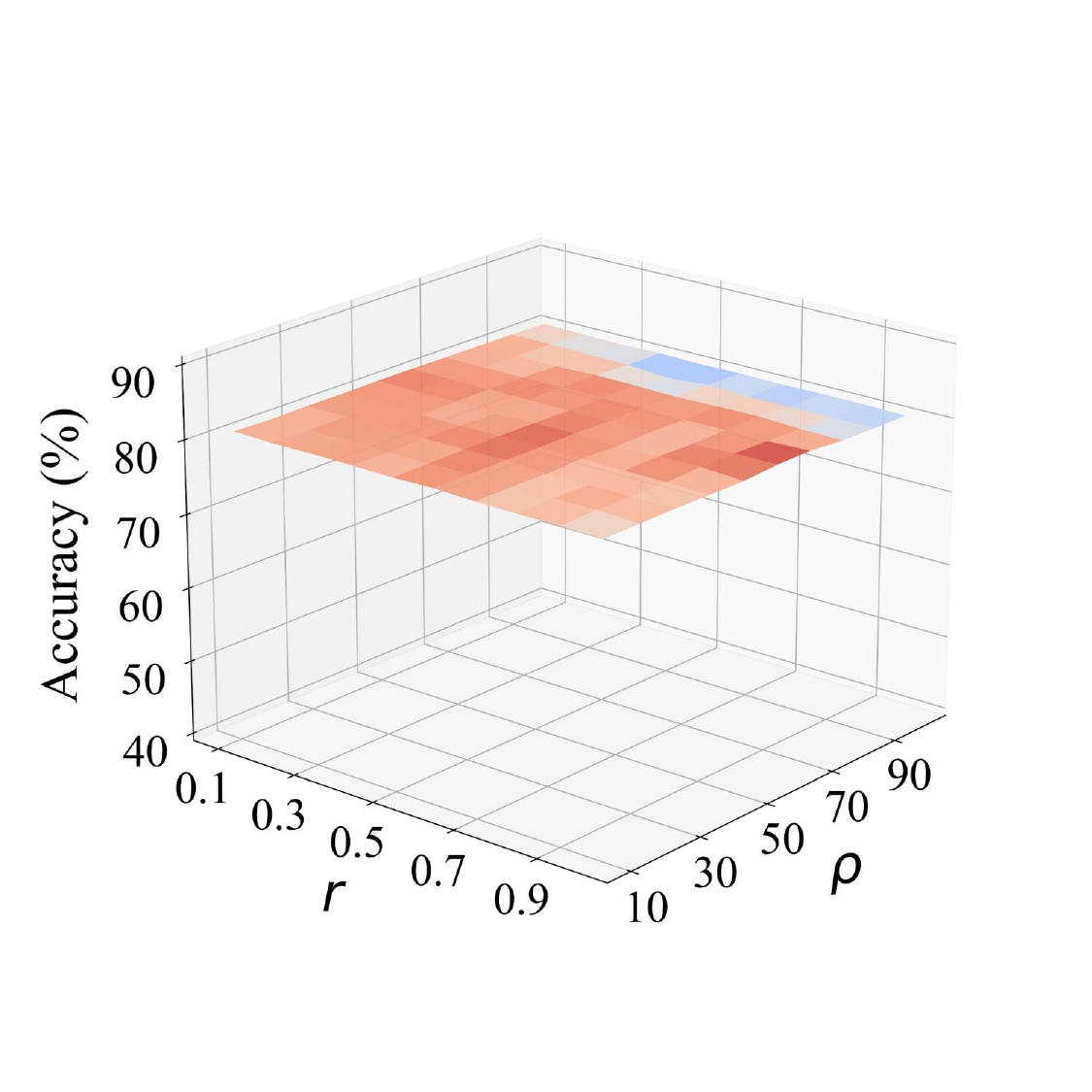}
    \end{minipage}%
    }%
    \vspace{-10pt}
    \caption{Parameter sensitivity analysis of $r$ and $\rho$ in our method at 60\% noise rate.}
    \label{fig:r_rho}
\end{figure*}

\begin{figure*}[t]
    \centering
    \subfigure[Cora]{
    \begin{minipage}[t]{0.15\linewidth}
    \centering
    \includegraphics[width=\linewidth, trim = 20 80 30 120, clip]{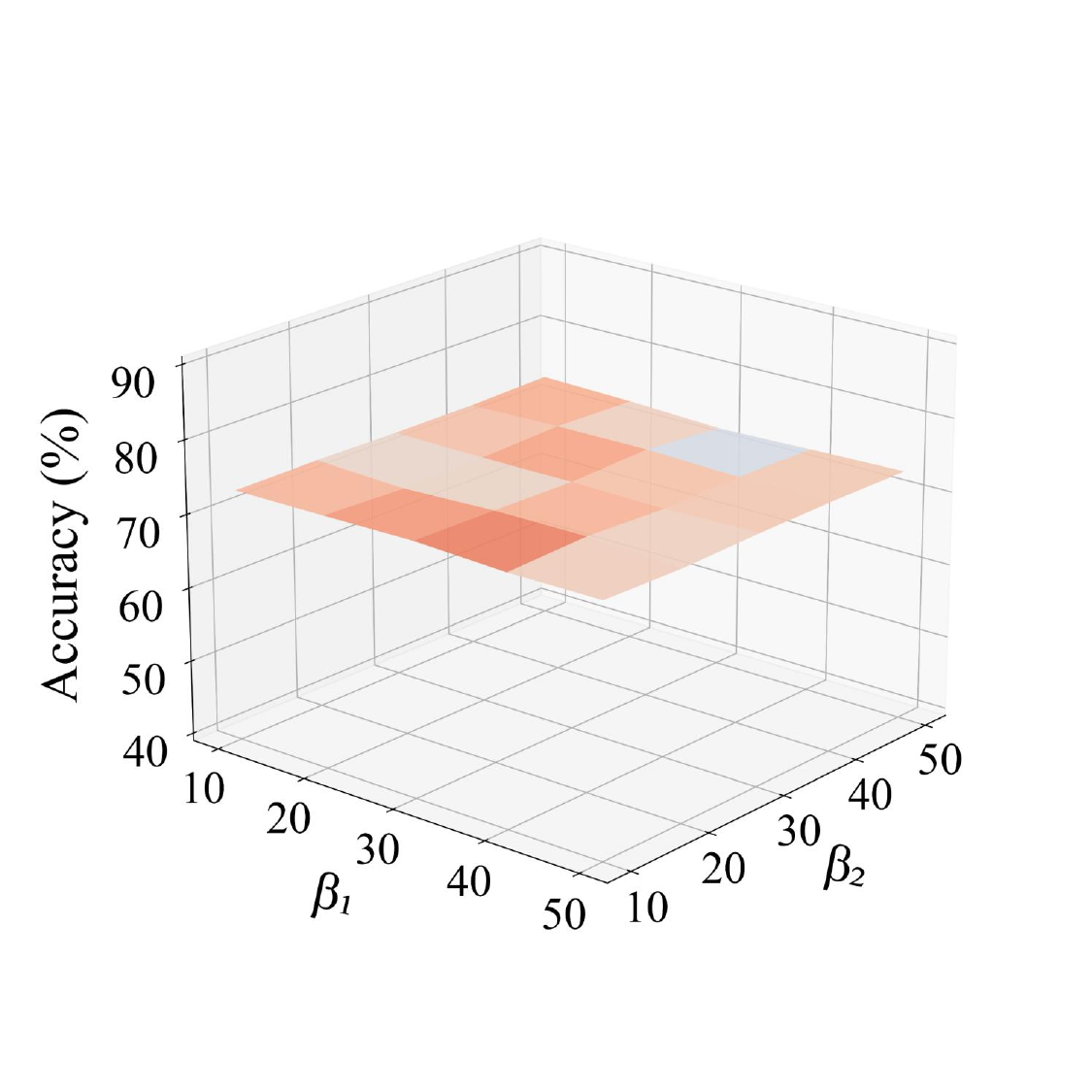}
    \end{minipage}%
    }%
    \hspace{0.01\textwidth}
    \subfigure[Citeseer]{
    \begin{minipage}[t]{0.15\linewidth}
    \centering
    \includegraphics[width=\linewidth, trim = 20 80 30 120, clip]{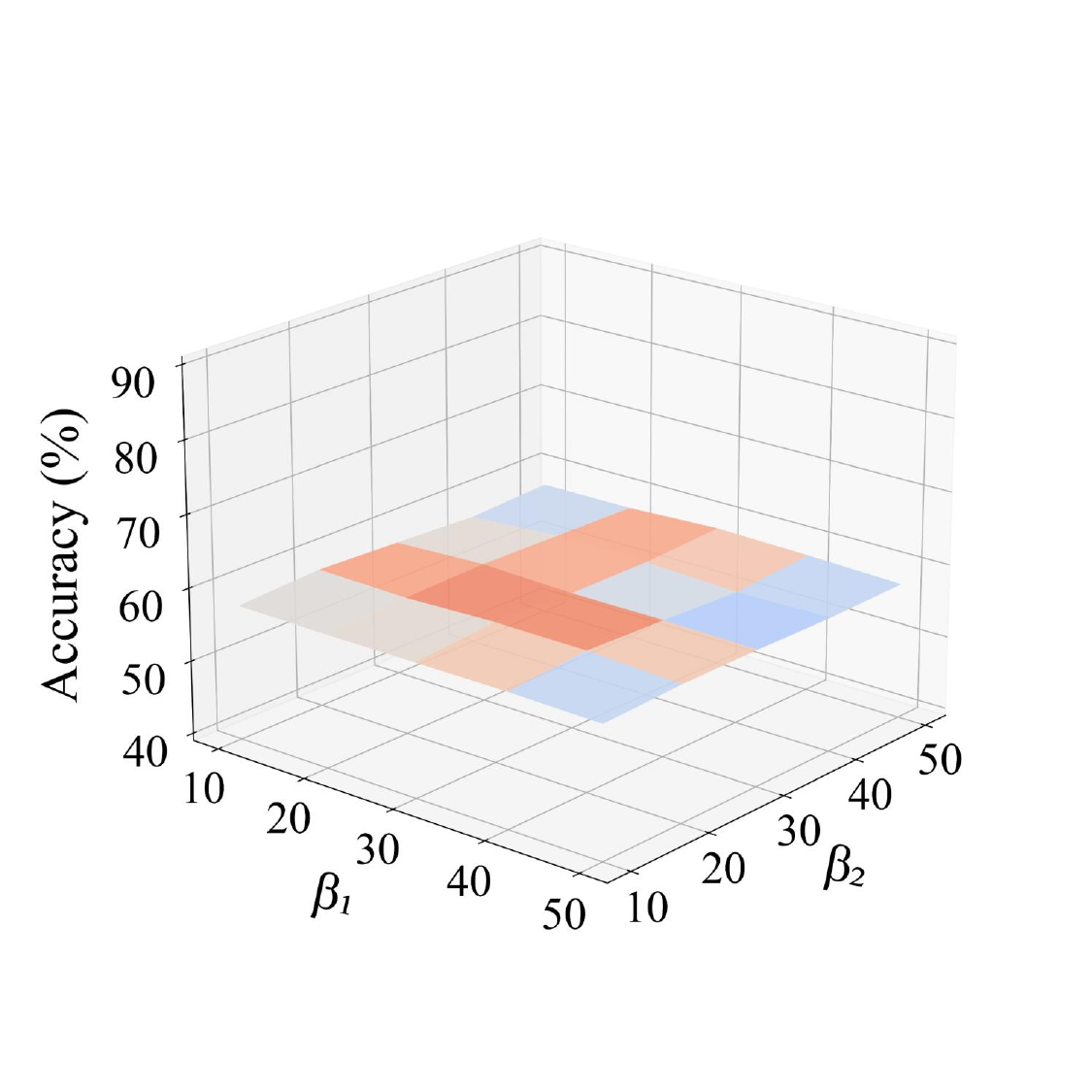}
    \end{minipage}%
    }%
    \hspace{0.01\textwidth}
    \subfigure[DBLP]{
    \begin{minipage}[t]{0.15\linewidth}
    \centering
    \includegraphics[width=\linewidth, trim = 20 80 30 120, clip]{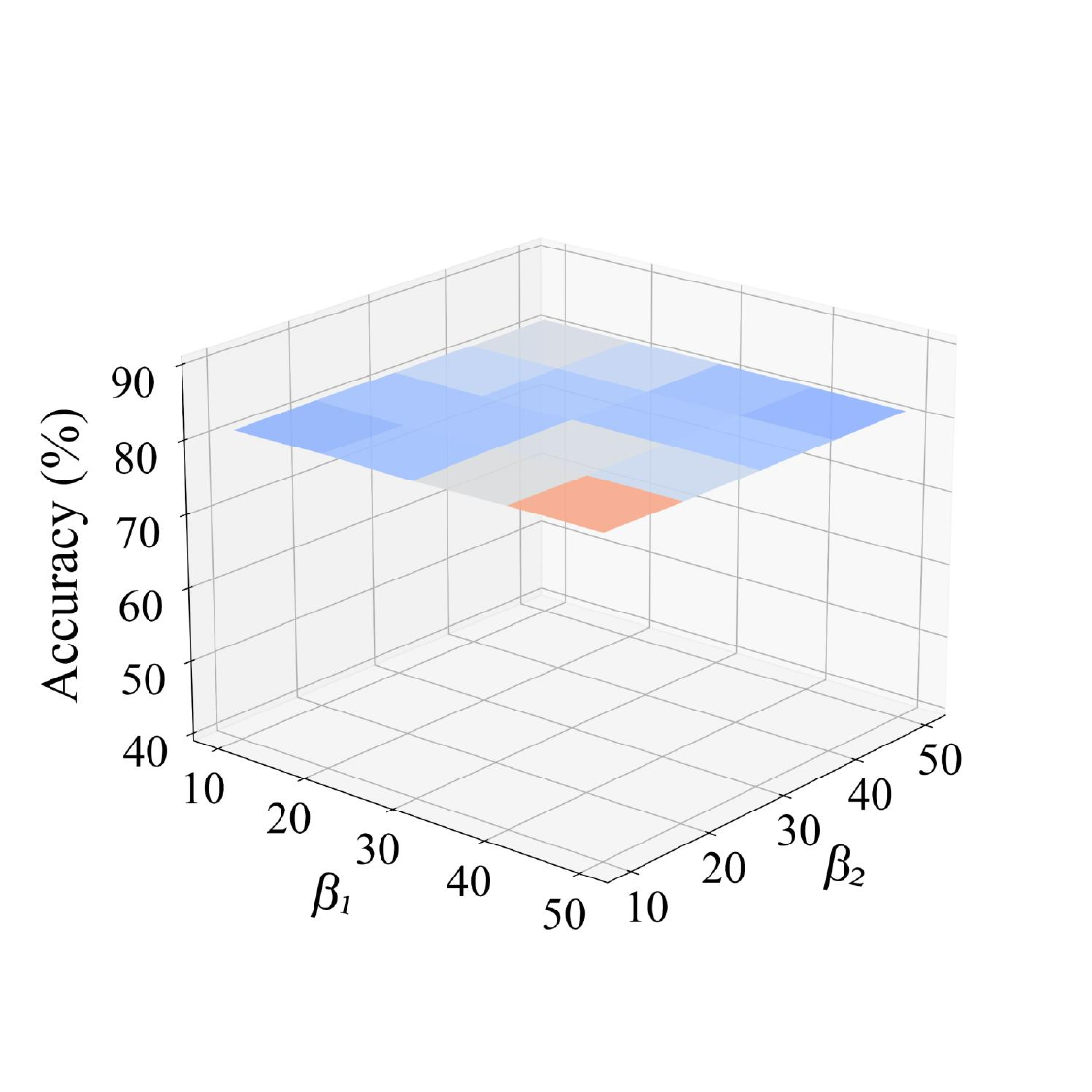}
    \end{minipage}%
    }%
    \hspace{0.01\textwidth}
    \subfigure[Photo]{
    \begin{minipage}[t]{0.15\linewidth}
    \centering
    \includegraphics[width=\linewidth, trim = 20 80 30 120, clip]{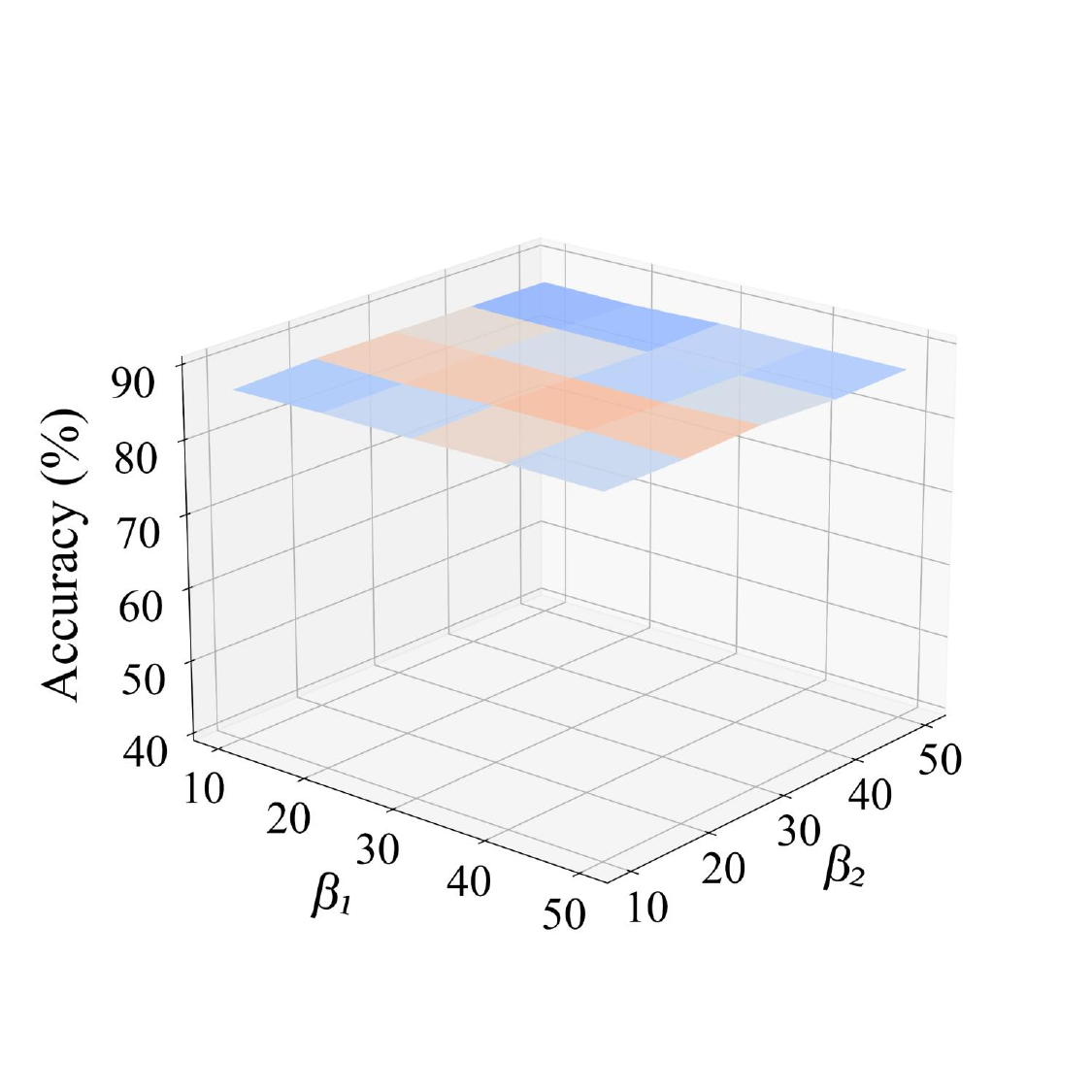}
    \end{minipage}%
    }%
    \hspace{0.01\textwidth}
    \subfigure[Computers]{
    \begin{minipage}[t]{0.15\linewidth}
    \centering
    \includegraphics[width=\linewidth, trim = 20 80 30 120, clip]{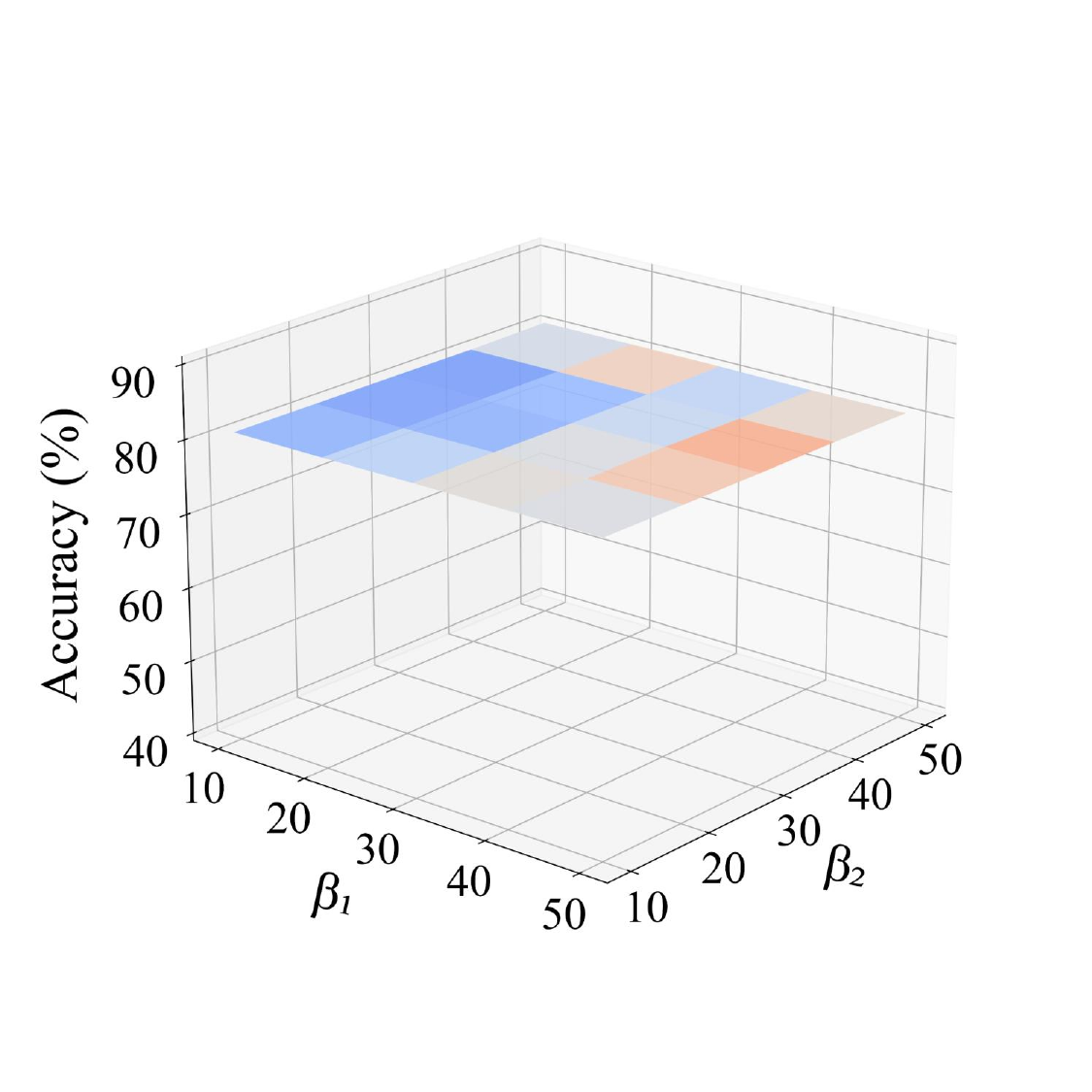}
    \end{minipage}%
    }%
    \vspace{-10pt}
    \caption{Parameter sensitivity analysis of $\beta_1$ and $\beta_2$ in our method at 60\% noise rate.}
    \label{fig:beta1_2}
\end{figure*}

\subsection{Ablation Study}
\label{section: Ablation Study}
Our method contains three important components, \ie using the student model to integrate knowledge from multiple teacher models (C1 for short), multi-teacher knowledge distillation based on bi-level optimization (C2 for short), and label improvement (C3 for short).
To demonstrate the effectiveness of each component, we report  the node classification performance of different component combinations on all datasets at the highest noisy rates in Table \ref{tab:ablation}. 


First, our method with all components averagely improves by 5.01\%, compared with the methods with one component only. This indicates that all three components are essential for our method, which verifies the feasibility of our proposed method. 
Second, the overall effectiveness of the method considering the student model is significantly superior to the one without considering the student model, because the student model can explore the complementary information from the teacher models.
Last,  bi-level optimization needs enough clean node to achieve effectiveness, while label improvement  may provide enough clean data. Hence, it is reasonable to simultaneously consider bi-level optimization based multi-teacher  distillation  and label improvement for dealing with noisy labels.

\begin{figure}[t]
    \centering
    \includegraphics[width=0.37\linewidth]{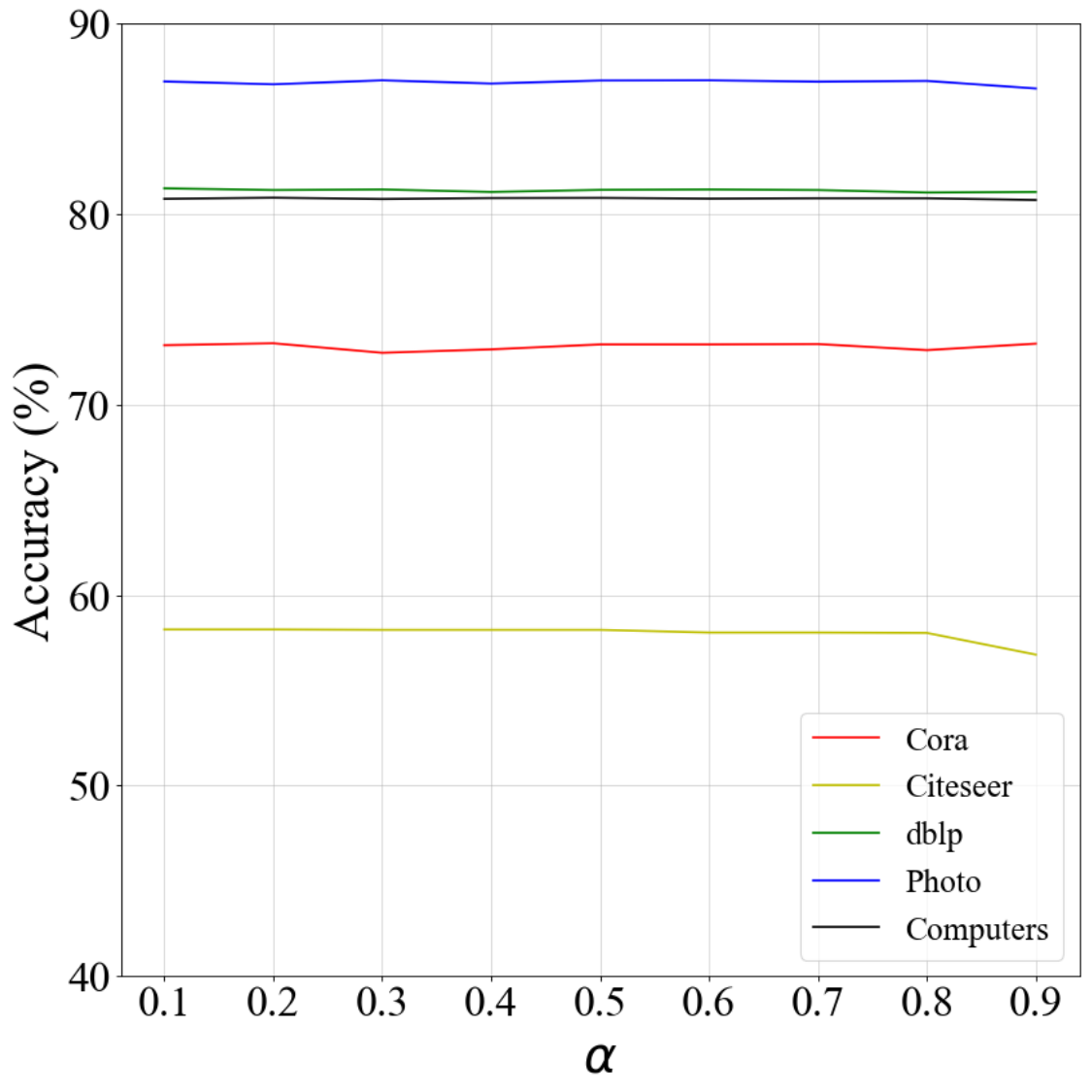}
    \caption{Parameter sensitivity analysis of $\alpha$ in our method at 60\% noise rate.}
    \label{fig:alpha}
\end{figure}

\subsection{Effectiveness of Multi-Teacher Construction}
We conduct  experiments to investigate the effectiveness of our constructed multi-teacher models on all datasets at the highest noisy rates  (\ie 60\%) for two types of noise and report the results in Appendix C.

Obviously, the performance of the student model increases with the increase of the number of teacher models. Specifically, compared to the best performance of the methods with only one teacher model, the results of our method with  three teacher models improves by an average of 2.68\% and 2.49\%, respectively, for uniform noise and pair noise  on all datasets. This suggests that multi-teacher models provide diverse complementary information to the student models.

\subsection{Effectiveness of Bi-level Optimization}
\label{Bi-level optimization is useful}

We compare our method with the mean method and the co-attention method on all datasets at the highest noisy rates (\ie 60\%) for two noise types and report the results in Table \ref{tab:Bo_useful}, to verify the effectiveness of the proposed bi-level optimization strategy.

Obviously, our proposed method achieves the best results. Specifically, compared with the best comparison method, \ie the co-attention method, our method improves by on average of 0.86\% and 1.88\%, respectively, for uniform noise and pair noise on all datasets. This fully demonstrates that our proposed method can more effectively capture the complementary information among teacher models, compared to all comparison methods.

\subsection{Parameter Sensitivity Analysis}
Our method has five key hyper-parameters, \ie $\rho$ for pseudo-label selection, $r$ for noisy label selection, and $\beta_1$, $\beta_2$, and $\alpha$ for selecting the clean node set. We investigate the sensitivity of our method to these hyper-parameters at 60\% noisy rate, and report the results in Figures \ref{fig:r_rho} \textasciitilde\ \ref{fig:alpha}.

Based on the experimental results, first, our method is not sensitive to the hyper-parameters, including $\beta_1$, $\beta_2$, and $\alpha$, because  our bi-level optimization process does not rely on a large number of correct labels.
Second, our method is sensitive to the parameters $r$ and $\rho$. Specifically, if the value of $r$ is less than 0.4, the performance of the model gradually increases with the increase of $r$. If the value of $r$ is larger than 0.6, the performance of the model gradually decreases with the increase of $r$. The reason is that too many correct labels were removed. , If the value of $\rho$ is less than 70, the variation of $\rho$ has little impact on the model performance. If the value of $\rho$ is larger than 70, the performance of the model gradually decreases with the increase of $\rho$. The reason is that the incorrect pseudo-label will  be inevitably selected.

\section{Conclusion}
In this paper, we proposed a new NNC method to address the limitations in previous methods. Specifically,  
multi-teacher distillations transfer diverse information to the learning of the student model. The bi-level optimization strategy fully explores the complementary information among multiple teacher models for the learning of the student model so that requiring less number of correct labels to achieve a robust student model. 
Experimental results showed that our method achieves supreme performance, compared to the state-of-the-art methods, in terms of noisy node classification tasks.

\clearpage

\bibliographystyle{Reference-Format}
\bibliography{Reference}

\appendix

\end{document}